\documentclass[summary]{URSIGASS2021}

\usepackage{algpseudocode}  
\usepackage{algorithmicx} 

\usepackage{amsmath}
\usepackage{amsfonts}       

\newcommand{\A}{{\mathcal{A}}}

\newcommand{\be}{\begin{eqnarray}}
  \newcommand{\ee}{\end{eqnarray}}
\newcommand{\beq}{\begin{equation}\begin{aligned}}
  \newcommand{\eeq}{\end{aligned}\end{equation}}
\newcommand{\beqn}{\begin{equation*}\begin{aligned}}
  \newcommand{\eeqn}{\end{aligned}\end{equation*}}
\newcommand{\ben}{\begin{eqnarray*}}
  \newcommand{\een}{\end{eqnarray*}}
\usepackage{algorithm}
\usepackage{multirow}
\usepackage{booktabs}       
\usepackage{colortbl}
\title{Low-Interception Waveform: To Prevent the Recognition of \\ Spectrum Waveform Modulation via Adversarial Examples}

\author{Haidong Xie$^{\dagger}$\affref{ref1}, Jia Tan$^{\dagger}$\affreftwo{ref1}{ref2}, Xiaoying Zhang\affref{ref1}, Nan Ji\affref{ref1}, Haihua Liao\affref{ref1}, Zuguo Yu\affref{ref2}\\
Xueshuang Xiang*\affref{ref1}, Naijin Liu*\affref{ref1}}
\affiliation{%
  \aff{ref1}{Qian Xuesen Laboratory of Space Technology, China Academy of Space Technology, China
  }
  \aff{ref2}{School of Mathematics and Computational Science, Xiangtan University, China}
  \affthank{$^\dagger$}{These authors contributed equally to this work.}\\
  \affthank{*}{Email: \{xiangxueshuang, liunaijin\}@qxslab.cn}
}

\begin{document}

\maketitle

\begin{abstract}
 Deep learning is applied to many complex tasks in the field of wireless communication, such as modulation recognition of spectrum waveforms, because of its convenience and efficiency. This leads to the problem of a malicious third party using a deep learning model to easily recognize the modulation format of the transmitted waveform. Some existing works address this problem directly using the concept of adversarial examples in the image domain without fully considering the characteristics of the waveform transmission in the physical world. Therefore, we propose a low-intercept waveform~(LIW) generation method that can reduce the probability of the modulation being recognized by a third party without affecting the reliable communication of the friendly party. Our LIW exhibits significant low-interception performance even in the physical hardware experiment, decreasing the accuracy of the state of the art model to approximately $15\%$ with small perturbations.
\end{abstract}

\section{Introduction}

With the development of the Internet of Things~(IoT), over $100$ billion devices are expected to be deployed in the IoT in the near future, leading to record-high requirements for wireless communication\cite{jagannath_machine_2019}. 
Deep learning~(DL) provides a general framework without predefined expert-selected features for solving complex tasks, such as automatic modulation recognition, 
and shows great benefit in wireless communication.
For example, O'Shea et al.\cite{oshea_over_2018} proposed a DL modulation recognition model with up to $94\%$ accuracy recently, greatly surpassing the traditional detectors. 
However, DL technology is a double-edged sword that  has revolutionized the industry while also opening the back door to malicious use.
Therefore, the security of DL models in wireless communication for non-cooperative games with third-party interventions is an important research topic\cite{pajola_threat_2019}.

\begin{figure}[t]
  \centering
  \includegraphics[width=\linewidth]{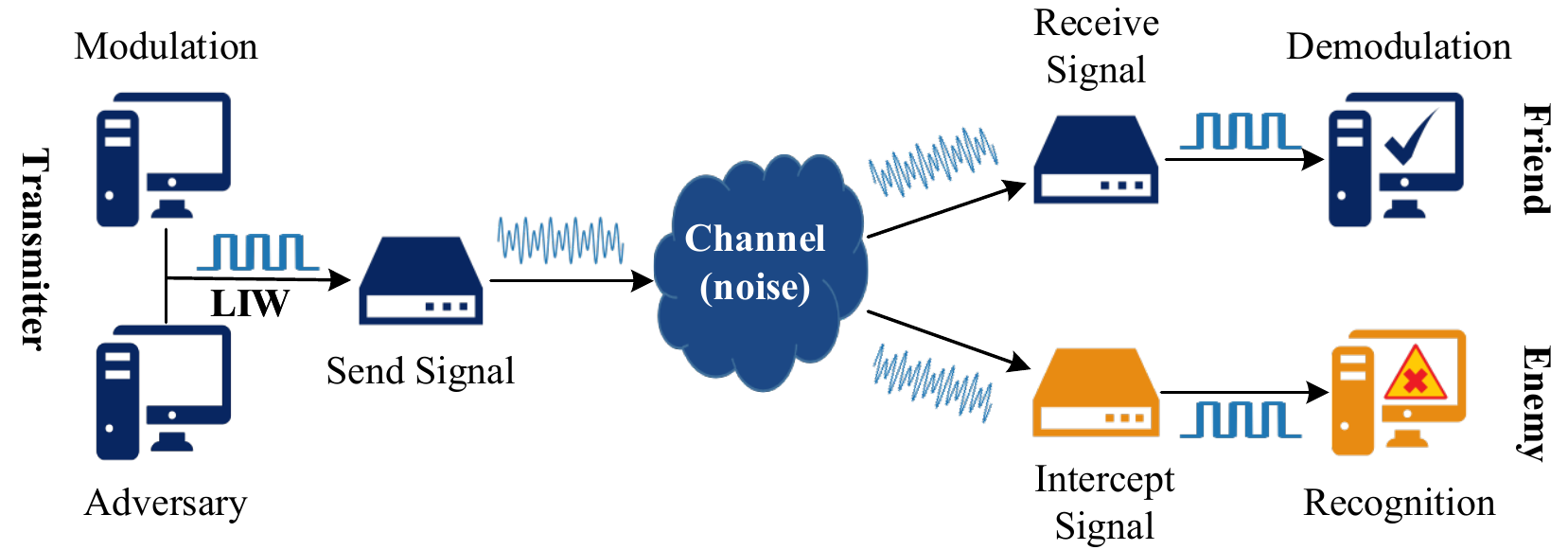}
  \vspace{-7mm}
  \caption{Scenario of wireless communication with Transmitter~(modulate and add adversary to generate LIW), Channel~(bring channel noise), Friend~(receive and demodulate signal) and Enemy~(intercept signal and recognize its modulation). Adding adversary but ignoring channel noise is ideal conditions, while practical conditions must consider the channel noise and hardware impact.}
  \label{fig:Sketch_Map}
\end{figure}

Let us consider a scenario of wireless communication with interception risk in Figure~\ref{fig:Sketch_Map}. 
To prevent the enemy from intercepting the signal by using DL models and ensure reliable communication between the transmitter and the friend, Sadeghi et al.\cite{sadeghi_adversarial_2019} first introduce adversarial examples~(AEs) into waveform modulation recognition, using the signal-based strategy\cite{8969541} to reduce the probability of modulation recognition by the enemy. 
The AEs above fool the model into outputting the wrong results by adding small and well-designed perturbations to original data\cite{szegedy_intriguing_2014,biggio_evasion_2013}. The study of Sadeghi et al.\cite{sadeghi_adversarial_2019} indicates that DL models in waveform modulation recognition tasks are not robust. 
Since then, many studies\cite{bair_limitations_2019,kokalj-filipovic_targeted_2019,restuccia_generalized_2020,kim_over--air_2020,delvecchio_investigating_2020} have introduced different AEs generation methods~(FGSM, PGD, C\&W) and effectively reduced the modulation recognition accuracy through a variety of different perspectives~(untargeted/targeted, white/black-box tasks).

Most of the above methods only apply the idea of AEs directly to the modulation recognition, and the waveforms that they obtain are generally based on $\ell_\infty$ norm with relatively large waveform perturbations that may not ensure reliable communication. At the same time, most reports in the literature do not consider the channel noise in practical conditions and rarely examine the effects of the waveform on a physical hardware platform. 
Therefore, this paper proposes \textbf{low-interception waveform (LIW)}, using the idea of decoupling direction and norm~(DDN) AEs generation method\cite{rony_decoupling_2019}. 
The LIW has the characteristics of low-interception performance~(strong attack capability) and small perturbation~(minimum adversarial perturbation with $\ell_2$ norm) to improve the security of the waveform with the minimal energy cost. 
Furthermore, considering the presence of a variety of complex channel noises and hardware quantization in practical conditions, we amplify the LIW perturbation and adjust the number of iterations to strengthen the suitability of LIW for practical application.

We verify the low-interception performance of LIW on the state of the art~(SOTA) model and the typical datasets\cite{oshea_over_2018} in Section~\ref{sec:result} both in ideal conditions and practical conditions based on the scenario in Figure~\ref{fig:Sketch_Map}. Experimental results show that LIW is effective in reducing the interception probability from $94\%$ of the original data to almost $0\%$ with a perturbation-to-signal ratio~(PSR\cite{sadeghi_adversarial_2019}) of only approximately $-20$dB~($1\%$ perturbation) in ideal conditions. 
Hardware platform experiments for practical conditions show that the probability of LIW being intercepted with higher SNR channel noise decreases to approximately $15\%$ by the addition of less than $-10$dB PSR perturbation. 

\section{Low-Interception Waveform}
\label{sec:liw}

As mentioned earlier, LIW refers to the DDN method\cite{rony_decoupling_2019} for the generation of the adversarial waveform that can be summarized as the following optimization equation: 
$\max_{\delta} \{\!\mathcal{L}(\theta, \mathbf{x}\!\!+\!\!\delta(\mathbf{x}),\mathbf{y}) \!-\! |\!|\delta(\mathbf{x})|\!|_2 \},$
where $\mathcal{L}(\theta,\mathbf{x}\!\!+\!\!\delta(\mathbf{x}),\mathbf{y})$ is the loss-function of the model, and $\delta(\mathbf{x})$ is the adversarial perturbation. Therefore, this equation is used to optimize AEs satisfying low-interception performance ($\mathcal{L}(\theta,\mathbf{x}\!\!+\!\!\delta(\mathbf{x}),\mathbf{y})$) and small perturbation ($|\!|\delta(\mathbf{x})|\!|_2$). 

\begin{algorithm}[h]
\caption{Low-Interception Waveform (LIW)}\label{alg:liw}
{\bf Input:} {Original data $x(n_0)$, true label or targeted label $y$, iteration number $K$, step size $\alpha$, norm modify factor $\gamma$ and perturbation scaling multiplier $\beta$.}\\{\bf Output:} {LIW $\tilde{x}$.}
\begin{algorithmic}[noend]
\State {Initialize: $\delta_0 \gets 0, \, \tilde{x}_0 \gets x(n_0), \, \epsilon_0 \gets 1$, \, \\ \quad\, m $\gets1\text{or}-1$~(non-targeted or targeted),}
\For {$k$ $\gets$ 1 to $K$} 
\State {$g \gets m\nabla_{\tilde{x}_{k-1}} \mathcal{L}(\theta,\tilde{x}_{k-1},y)$,} 
\Comment{Calculate gradient}
\State {$\delta_k \gets \delta_{k-1} + \alpha\frac{g}{\left \| g \right \|_2}$,}
\Comment{Update direction}
\State \textbf{if} {$\tilde{x}_{k-1}$ is adversarial,} \textbf{then} {$\epsilon_k \gets (1-\gamma) \epsilon_{k-1}$,} 
\State \textbf{else} {$\epsilon_k \gets (1+\gamma)\epsilon_{k-1}$,}
\Comment{Reduce or enlarge norm}
\State $\tilde{x}_k \gets clip(x + \epsilon_k\frac{\delta_k}{\left \| \delta_k \right \|_2},0,1) $, \Comment{Update and clip LIW}
\EndFor
\State {$\tilde{x} = x + \beta(\tilde{x}_K - x)$.} \Comment{Enhance LIW by multiplier}
\end{algorithmic}
\end{algorithm}

The full procedure of generating LIW is described in Algorithm~\ref{alg:liw}. We start from the original data $x(n_0)$ and iteratively refine direction $\delta_k$ using the current and historical gradient, and either reduce or enlarge the perturbation norm $\epsilon_{k} = (1 \mp \gamma)\epsilon_{k-1}$ if the current waveform $\tilde{x}_{k-1}$ either is or is not adversarial, and then update LIW $\tilde{x}_k$ of each iteration $k$. After $K$ iterations, we obtain the LIW AEs $\tilde{x}_K$, and while considering the effect of channel noise in waveform propagation, we finally amplify the LIW $\tilde{x}$ by a factor $\beta$. Considering hardware quantization, the final LIW is quantified to $8$-bit for the practical case. 

In this work, the selection of $K (100 \text{or} 10)$ and $\beta (1 \text{or} 10)$ is described in detail in Section~\ref{sec:result}, and we set $\alpha$ from $1.0$ to $0.01$ with cosine annealing and $\gamma = 0.05$.
To obtain the minimum perturbation, we use the non-targeted algorithm, and the corresponding targeted algorithm can be applied to other scenarios; this is not discussed in this paper. 
Moreover, we examine the performance of LIW $\tilde{x}$ in the presence of original data noise~$n_0$ or channel noise~$n$ with different SNR. In ideal conditions, we directly validate the accuracy $\A(\tilde{x}(n_0))$, but in practical conditions, we validate the accuracy by $\A(\tilde{x}(n_0\!=\!30) \!+\! wgn(x,n))$, where $wgn(x,n)$ is Gaussian white noise with given channel noise intensity $n$. 

\section{Hardware Platform Practical Evaluation}
\label{sec:hard}

\begin{figure}[t]
  \centering
  \includegraphics[width=\linewidth]{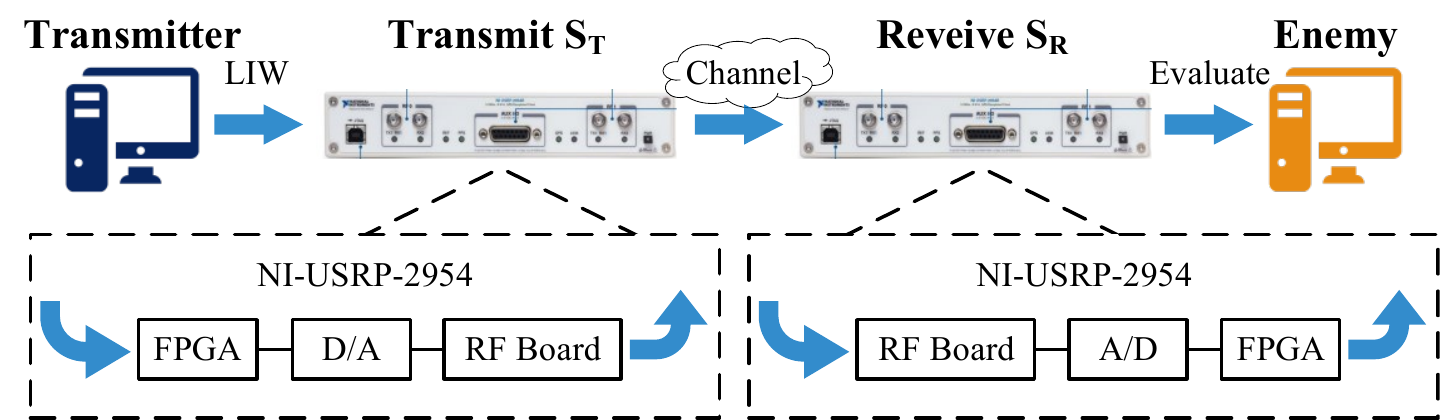}
  \vspace{-5mm}
  \caption{Illustration of hardware evaluation for LIW.}
  \label{fig:Hardware}
\end{figure}

For practical conditions, this work not only carries out numerical simulations, but also innovatively uses the hardware platform to carry out the verification, because the numerical results in and of themselves cannot fully represent the effects encountered in practical application.
Figure~\ref{fig:Hardware} \& Algorithm~\ref{alg:hard} provides a detailed introduction of the hardware spectrum signal transmitting and receiving platform built using NI-USRP-$2954$\cite{USRP,UHD} equipment and its evaluation process for verifying the performance of LIW encountering the practical channel.

\begin{algorithm}[h]
  \caption{LIW Hardware Platform Evaluation Process} 
  \label{alg:hard}  
  {\bf Set up:} Central frequency, sampling rate, signal gain and channel noise $n$ of signal transmission. \\
  {\bf Step~1:} Prepare original waveform dataset $\{x(n_0=30)\}$, (approximately considered that $n_0=30$ data is noise free,)\\
  {\bf Step~2:} Generate LIW and obtain dataset $\{\tilde{x}\}$, \\
  {\bf Step~3:} Splice $\{\tilde{x}\}$ into signal $S_T$ to be transmitted, \\
  {\bf Step~4:} Transmit the signal $S_T$ through the hardware platform and receive signal $S_R$, \\
  {\bf Step~5:} Split the received signal $S_R$ into dataset $\{\tilde{x}_R\}$, \\
  {\bf Step~6:} Evaluate the received dataset $\{\tilde{x}_R\}$.
\end{algorithm}

\section{Experiment Results}
\label{sec:result}

This section examines the low-interception performance of LIW on the SOTA ResNet model and the 2018.01.OSC dataset\cite{oshea_over_2018} both by using numerical simulations and hardware evaluation.
The dataset contains a total of up to $2555904$ data for $24$ different modulations and $26$ different original data SNR $n_0 \!\in\! [-20,30]$. 
All experiments are executed on a desktop computer with RTX 2080 GPU. Approximately $11$ hours are required to train the ResNet model for $50$ epochs, and approximately $0.035$ seconds are required to generate each LIW with $K=100$. 

\subsection{The SOTA Model}

\begin{figure}[t]
  \centering
  \vspace{-3mm}
  \includegraphics[width=\linewidth]{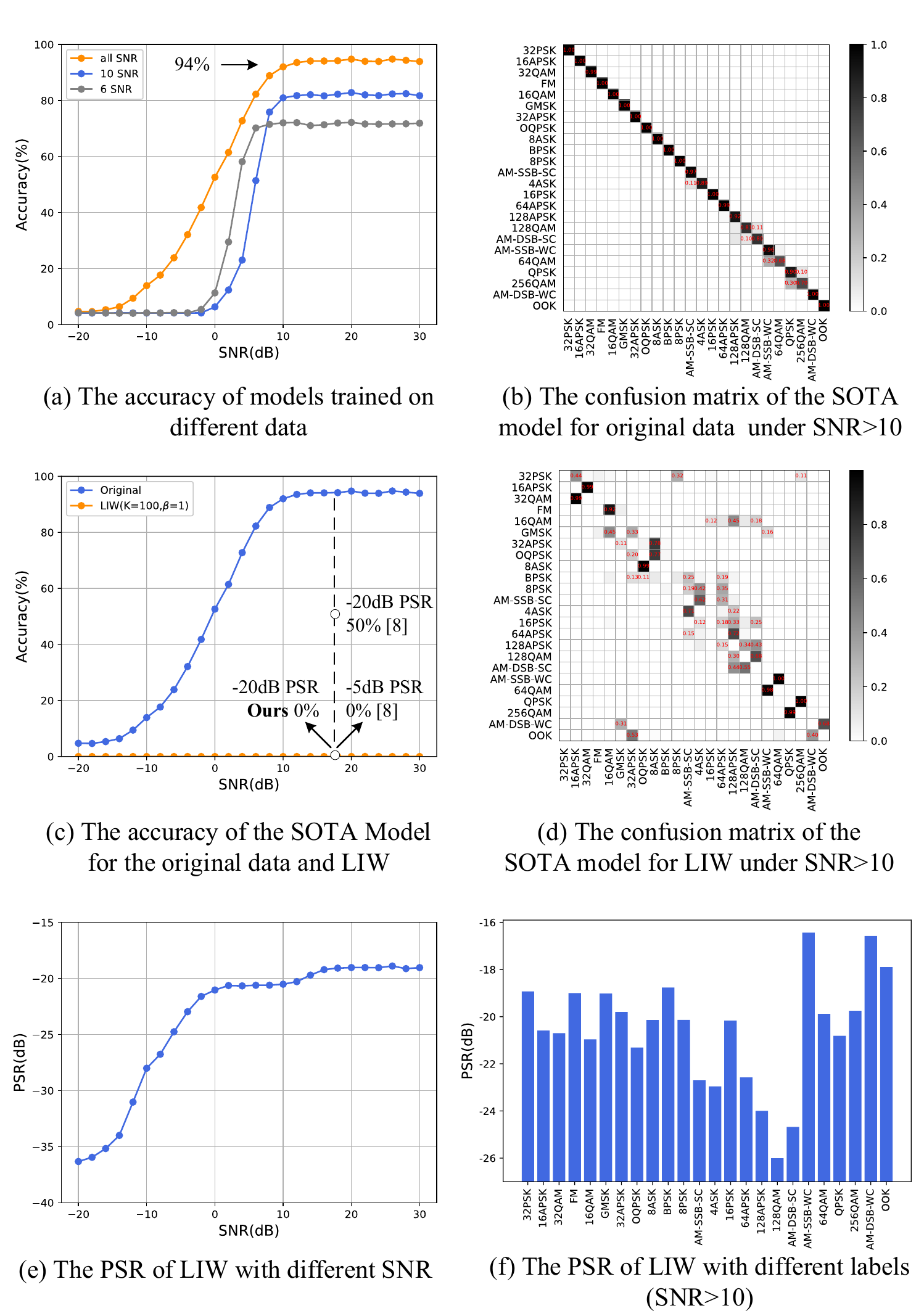}
  \vspace{-6mm}
  \caption{Effect of the model and LIW in ideal conditions.}
  \label{fig:SNRall}
\end{figure}

First, considering only the influence of noise on modulation recognition, we reproduce the SOTA ResNet model in Reference\cite{oshea_over_2018}. We find that there exists an intrinsic correlation between the different SNR data. The model trained with mixed all SNR data has a higher accuracy than any single SNR data, reaching approximately $94\%$ as shown in Figure~\ref{fig:SNRall}(a), and its confusion matrix is shown in Figure~\ref{fig:SNRall}(b).
Overall, the accuracy of the model trained with the data increases gradually with increasing SNR.

\subsection{LIW in Ideal Conditions }

To fully reflect the performance in ideal conditions, i.e., without considering channel noise, we generate LIW with the number of iterations $K\!=\!100$ without perturbation scaling. It is clear from Figure~\ref{fig:SNRall}(c) that the recognition accuracy of the model can be reduced to almost $0\%$ by LIW, indicating that it can almost completely prevent the recognition of waveform modulation. 
Figure~\ref{fig:SNRall}(d) shows the confusion matrix of LIW for SNR $\!>\!10$ and displays more details regarding the LIW's excellent low-intercept performance. 
On the other hand, it is clear that the difference between LIW and original data are very small as indicated in Figures~\ref{fig:SNRall}(e and f). The overall PSR of only approximately $-20$dB ($1\%$ perturbation) that is difficult to distinguish by human eye, shows that LIW has little effect on reliable communication between the transmitter and the friend. 

By contrast, the best current waveform in the literature shown in Figure~\ref{fig:SNRall}(c) requires a PSR as high as $-5$dB to achieve approximately $0\%$ accuracy, while it can only achieve $50\%$ accuracy at $-20$dB\cite{bair_limitations_2019}. Although they use AEs based on $\ell_\infty$ or other norm can achieve good effect, the corresponding perturbation is relatively large. We use the DDN method based on $\ell_2$ for LIW to obtain the minimum perturbation, therefore \textbf{LIW has better low-interception performance without affecting reliable communication.}

\subsection{LIW in Practical Simulation}

\begin{figure*}[t]
  \centering
  \vspace{-3mm}
  \includegraphics[width=\linewidth]{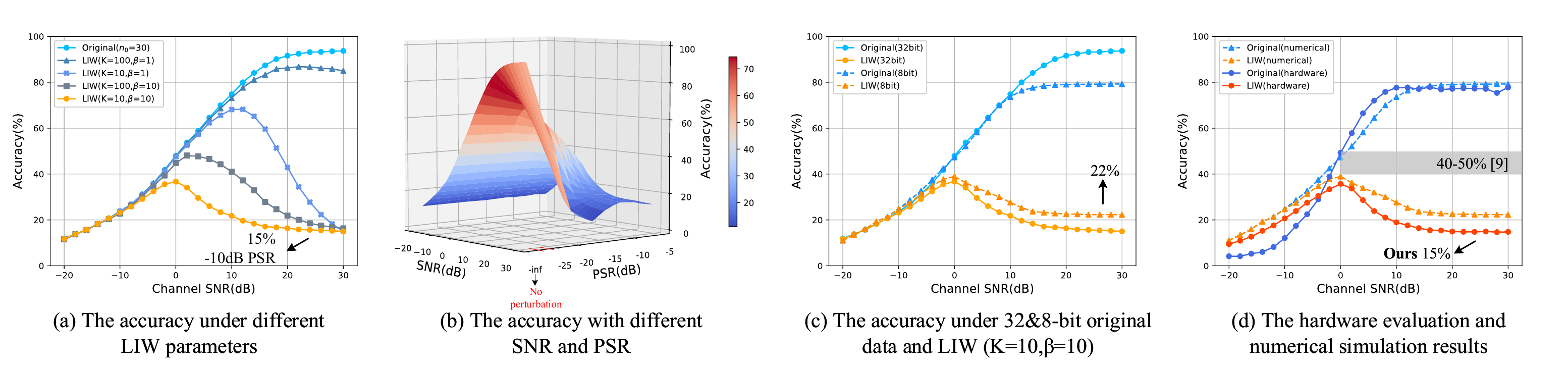}
  \vspace{-7mm}
  \caption{Performance of LIW in practical conditions and hardware evaluation.}
  \label{fig:practice}
\end{figure*}

In practical conditions, LIW should be generated on the basis of clean data(SNR=30) and take into account the effects of channel noise and hardware quantization. 
We examine different LIW parameter strategies and find that number of iterations $K$ and perturbation scaling multiplier $\beta$ are important when facing unknown channel noise during transmission. Although LIW with $K=100$ has excellent low-interception performance in ideal conditions, it is susceptible to interference from channel noise. We believe that this is because the added perturbation is too specific to have generalization ability and the too small perturbation is easily covered by channel noise.

As shown in Figure~\ref{fig:practice}(a), the experiments validate the above idea. Reducing the number of iterations K to 10 enhances the generalization ability of LIW without increasing the perturbation, and the best low-interception performance of LIW is obtained by scaling up the waveform 10 times on this basis. Such an approach successfully decreases the model accuracy under channel noise with multiple SNRs to approximately $15\%$, and the PSR of the added perturbation is only approximately -10~dB.
To better observe the three-way relationship between the channel noise, added perturbation and low-interception accuracy, we plot it as a 3D graph as shown in Figure~\ref{fig:practice}(b). It is concluded that \textbf{when the SNR of channel noise is large, the required perturbation is smaller, but when it is small, a larger perturbation must be added to achieve good low-interception accuracy.}

In the ideal numerical experiment, we use $32$-bit data in quantization, but in practical conditions the transceiver hardware usually can only send and receive low-quantization signal such as $8$-bit.
Therefore, we convert the $32$-bit original data and LIW to $8$-bit and add channel noise to simulate practical conditions. The recognition accuracy of the original data decreases to a certain extent from $94\%$ to approximately $80\%$. The corresponding LIW with $K=10$ and $\beta=10$ shows the low-interception probability from $15\%$ to approximately $22\%$ as shown in detail in Figure~\ref{fig:practice}(c). Therefore, quantization will have a certain impact on LIW, but the impact is weak.

\subsection{LIW in Hardware Evaluation}

The last set of experiments are carried out to evaluation whether the low-interception performance of LIW is maintained when it is used on a hardware platform.
The results presented in Figure~\ref{fig:practice}(d) show that the accuracy curve obtained in the hardware evaluation process is basically consistent with the trend of the numerical simulation data. The original accuracy is still high at higher SNR with approximately $80\%$, but drops sharply for lower SNR. LIW has a relatively stable low-interception performance particularly at higher SNR of approximately $15\%$. 
By contrast, similar experiments in the previous literature could only reduce the recognition accuracy to $40-50\%$\cite{kokalj-filipovic_targeted_2019}. 
These results fully show that the factors considered here can cover practical conditions to a great extent, so that \textbf{LIW is suitable for application in practical physical devices}. 

\section{Conclusion}

To avoid the spectrum waveform modulation being malicious recognized by DL models, this paper proposes a LIW method to lower the risk of interception. The core idea of LIW is to introduce DDN with the specially designed parameters together with amplification scaling of the generated LIW perturbation, so that LIW with only minor PSR can avoid being intercepted by the recognition model. 
Based on experiments with ideal conditions, practical conditions and on a hardware platform, we conclude that the LIW shows outstanding low-interception performance in both numerical and physical experiments and has strong application potential. 
Of course, the work described here represents only the initial exploration of LIW, and many physical problems must still be solved in future work. 

\section{Acknowledgements}

This work was supported by the Innovation Foundation of Qian Xuesen Laboratory of Space Technology.

\small
\bibliographystyle{ieeetr}
\bibliography{paper}

\end{document}